%% file: 00-cikm21.tex
\documentclass[sigconf]{acmart}
\usepackage{multicol}
\usepackage{graphicx}
\usepackage{colortbl}

\def\BibTeX{{\rm B\kern-.05em{\sc i\kern-.025em b}\kern-.08emT\kern-.1667em\lower.7ex\hbox{E}\kern-.125emX}}
    
\copyrightyear{2021} 
\acmYear{2021} 
\setcopyright{acmcopyright}\acmConference[CIKM '21]{Proceedings of the 30th ACM International Conference on Information and Knowledge Management}{November 1--5, 2021}{Virtual Event, QLD, Australia} \acmBooktitle{Proceedings of the 30th ACM International Conference on Information and Knowledge Management (CIKM '21), November 1--5, 2021, Virtual Event, QLD, Australia} 
\acmPrice{15.00} 
\acmDOI{10.1145/3459637.3482376}
\acmISBN{978-1-4503-8446-9/21/11}

\begin{document}

\title{POSHAN: Cardinal POS Pattern Guided Attention for News Headline Incongruence}

\author{Rahul Mishra}
\affiliation{University of Stavanger, Norway}
\email{rahul.mishra@uis.no}
\author{Shuo Zhang}
\affiliation{Bloomberg, United Kingdom}
\email{szhang611@bloomberg.net}

\begin{abstract}
Automatic detection of click-bait and incongruent news headlines is crucial to maintaining the reliability of the Web and has raised much research attention. However, most existing methods perform poorly when news headlines contain contextually important cardinal values, such as a quantity or an amount. In this work, we focus on this particular case and propose a neural attention based solution, which uses a novel cardinal \textbf{Part of Speech (POS)} tag pattern based \textbf{h}ierarchical \textbf{a}ttention \textbf{n}etwork, namely \textbf{\emph{POSHAN}}, to learn effective representations of sentences in a news article. In addition, we investigate a novel cardinal phrase guided attention, which uses word embeddings of the contextually-important cardinal value and neighbouring words. In the experiments conducted on two publicly available datasets, we observe that the proposed methodgives appropriate significance to cardinal values and outperforms all the baselines. An ablation study of POSHAN shows that the cardinal POS-tag pattern-based hierarchical attention is very effective for the cases in which headlines contain cardinal values.

\end{abstract}


%

\maketitle

\input{01-introduction}
\balance
\input{02-relatedwork}

\input{03-problem}
\input{04-data}
\input{06-results}
\input{07-conclusion}

\balance

\bibliographystyle{ACM-Reference-Format}
\bibliography{references}
\end{document}

%% file: 01-introduction.tex
\section{Introduction}
\label{sec:intro}

\begin{figure}[t]
      	\centering
	\includegraphics[scale=0.54]{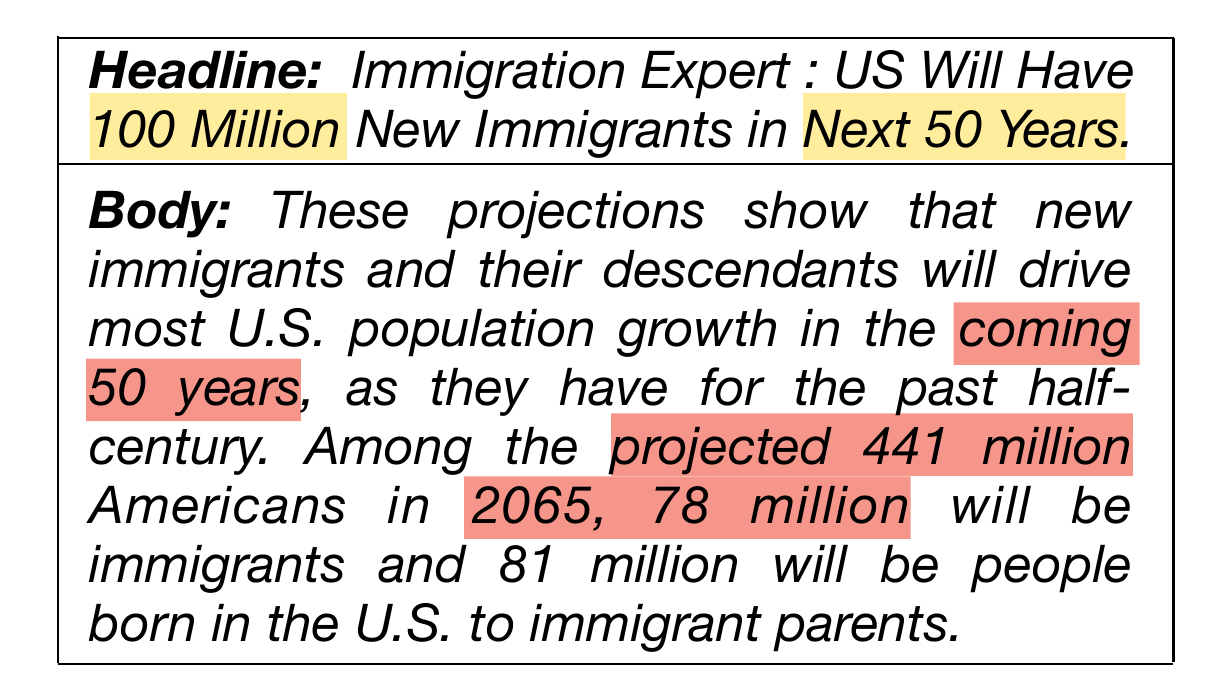}
      	\caption{Example of an Incongruent  Headline.  The headline says, in the next 50 years, there will be 100 million new immigrants but the news body quotes about only 78 million new immigrants. }
      	\label{fig:example}
      		  \vspace*{-17pt}
\end{figure}

News titles expose the first impression to readers and decide the viral potential of news stories within social networks~\cite{Reis}.
Most of the users only rely on the news title to decide what to read further \cite{Gabielkov}.
A deceptive and misleading news title can lead to false beliefs and wrong opinions.   It becomes inversely worse when users share the news on social media without reading the news body but only skimming through the news title. The news headlines, which are ambiguous, misleading, and deliberately made catchy to lure the users to click, are called incongruent headlines or click baits \cite{Wei}. Figure~\ref{fig:example} illustrates an example.
There is a line of study that has been investigated and analyzed in the literature \cite{Yimin,7752207, Potthast,ferreira-vlachos-2016-emergent,ijcai2017-579,Wei,8594871,Yoon}, witnessed by different techniques such as linguistic feature based methods \cite{Yimin}, generative adversarial networks\cite{8594871, MuSem}, and hierarchical neural attention networks~\citep{Yoon}. 
For example, \citet{Yoon} propose a headline text guided neural attention network to compute an incongruence score between the news headline and the corresponding body text. They introduce a hierarchical attention based encoder, which encodes words of news body text at the word level to form paragraph representations and encodes paragraphs to form document representation.  
However, we observe that these prior works fail to generalize and perform adequately in cases where news headlines contain a significant numerical value. 
The numerical values can be in the form of a currency amount, counts of people, months, years or objects, etc. 
For instance, in Figure \ref{fig:example} an excerpt from a news item is shown, in which the news headline \emph{``Immigration Expert : US Will Have 100 Million New Immigrants in Next 50 Years.''}, contains two contextually important numerical figures i.e. \emph{``100 Million''},\emph{``50 Years.''}.  The headline mentions, there will be 100 million new immigrants, but the news body quotes only 78 million new immigrants. 
The headline is deliberately made contradictory and exaggerating to look more sensational. It's apparent from this example that numerical and cardinal values are useful and crucial cues of the congruence of the news headlines.

All of the prior works suffer from not giving enough importance to numerical values. The headline guided attention-based methods such as~\citep{Yoon}, fail to attend relevant words related to cardinal phrases as they do not treat them specifically. 
On the other hand, generative adversarial network-based methods, which generate a synthetic headline from news body text to augment the dataset or to use them for similarity matching with original headlines, also miss cardinal aspects in the synthetically generated headlines. 
Clearly, the news headlines having numbers are not trivial cases for incongruence detection, and in this paper, we try to deal with news headline incongruence detection with special focus to such cases. 

The objective of this work is to devise an incongruence detection method, which not only performs better than previously proposed techniques but also resolves the deterioration of classification accuracy with the news items in which the headline contains cardinal values.
In specific, we leverage a novel Cardinal Part-of-Speech Tag patterns to drive the hierarchical neural attention to capture salient and contextually important words and sentences at the word and sentence levels correspondingly.  
The key idea of using cardinal pos patterns such as  $(NN:CD:JJ)$ or $(VBD:CD:CD)$ is to use them as latent features associated with news headlines and learn the contextual embeddings based on data samples containing the same cardinal pos patterns. 
The embeddings are used at the test time to drive the attention and capture the salient words and sentences, which are significant for cardinal values. 
In addition, we investigate a cardinal phrase guided attention mechanism and combine both with standard headline guided attention.
To utilize the better contextual representation of words, we fine tune the pre-trained BERT model and extract the word embeddings, which are fed to a Bi-LSTM based sequence encoder.  
We conduct experiments with the subset of two publicly available datasets and achieve state-of-the-art performance. 
The proposed model \emph{POSHAN} not only outperforms all the other methods in original datasets but also its performance does not deteriorate much compared to other models, with derived datasets, containing only those data samples, which have numerical values in the news headlines.
We visualize the Cardinal POS Pattern embeddings and overall attention weights to further analyse the effectiveness of the proposed model. We observe that the Cardinal POS Patterns have formed clearly separated clusters in embedding space, which connote the congruence and incongruence labels. It is apparent from the visualization of overall attention weights, that  \emph{POSHAN} model successfully attends the contextually important cardinal phrases in addition to other significant words.
In nutshell, the major contributions of this work are: 
\begin{itemize}
\item We focus on the news headline incongruence detection when news headlines containing numbers, and propose the cardinal POS pattern guided attention (Section~\ref{cardinal pos pattern}) baseline.
\item We propose a cardinal phrase guided attention (Section~\ref{cardinal phrase}) mechanism and combine the both cardinal POS pattern and cardinal phrase attention with standard headline guided attention (Section~\ref{headline attention}) in a joint model (Section~\ref{fuse}).
\item We incorporate the proposed hierarchical attention methods on top of a Bi-LSTM based sequence encoder (Section~\ref{Sequence}) which encodes the sequence of fine-tuned (Section~\ref{implement}) pre-trained BERT embeddings (Section~\ref{embedding}) of the words.
\item In the evaluation with two publicly available datasets (Table~\ref{tab:nelaresult} and \ref{tab:CBCresult} ), the proposed techniques outperform the baselines and state-of-the-art methods. 
\item We visualize the Cardinal POS Pattern embeddings and overall attention weights and conduct error analysis to analyze the effectiveness of the proposed model, and verify the effectiveness.
\end{itemize}

%% file: 02-relatedwork.tex
\section{Related Work}
\label{sec:relatedwork}
Detection and prevention of misinformation and deceptive content online has gained lots of traction recently. Incongruent news and click-baits are very common forms of deception and misinformation. Naturally, most of the prior works in this area have treated the click-baits or news incongruence detection task as a standard text classification problem. Majority of the initial works are feature engineering heavy \cite{Biyani}, exploiting diverse features such as linguistic features, lexicons, sentiments and statistical features. 
\citet{7752207} use linguistic and syntactic features such as sentence structure, word patterns, word n-grams and part-of-speech (POS) n-grams etc. and learn a classifier using support vector machine (SVM) to detect click-baits.
\citet{Potthast} use text features and meta-information of tweets such as entity mentions, emotional polarity, tweet length and word n-grams to learn a classifier, experimenting with methods such as random forest, logistic regression etc.to detect click-baits. 
\citet{Yimin} propose to conduct lexical and syntactic analysis and advocate to utilize image features and user-behavior features for the identification of the click-baits. These methods are outperformed by the recent deep learning based methods\cite{Vaibhav, Ankesh}, in which hand crafted feature engineering is not required. 
News headline incongruence is closely related to a number of tasks such as sentence matching based stance classification~\cite{ferreira-vlachos-2016-emergent, ijcai2017-579}.
%
Sentence pair classification task using fine tuned pre-trained language models such as BERT \cite{devlin-etal-2019-bert} and RoBERTa \cite{liu2019roberta} has received a great traction from the community and it is a closely related problem to headline incongruence. Sentence pair classification typically consists a pair of sentences, while in headline incongruence systems, we need to deal with a sentence and a large news body content in order to form the evidence for congruence. 
The sentence matching and lexical similarity based methods \cite{neculoiu-etal-2016-learning} are not a good fit for headline incongruence problem due to inherent challenges such as relative length and vocabulary mismatch between the news headline text and its body content.
Therefore, these tasks share the advancement of the development of the techniques.
For example, 
\citet{Wei} introduce a co-training based approach with myriad kinds of features such as sentiments, textual, and informality. 
Recent works such as \cite{Yoon} use neural attention \cite{Bahdanau2015} based approach to achieve headline guided contextual representation of the news body text and also release a Korean and an English dataset for headline incongruence.

Although some recent works such as \cite{Yoon} have achieved state-of-the-art performance, most of the existing approaches do not perform well in the case of the headline containing cardinal numbers because no additional emphasis is given to the cardinal numbers.
\citet{8594871} use a generative approach to augment the dataset by additionally generating synthetic headlines. \citet{MuSem} use an inter-mutual attention-based semantic matching between the original and a synthetically generated headlines via generative adversarial network based techniques, which utilises the difference between all the pairs of word embeddings of words involved and computes mutual attention score matrix. These generative methods are also not very useful in the task at hand as the news headlines, generated using news body content, usually miss the cardinal information.
Focusing on the quantity cases, we propose a neural attention mechanism in which, we use novel cardinal POS triplet and cardinal phrase guided attention in addition to standard headline guided attention. This technique makes sure to have two contextual information: firstly, by using headline guided attention, all the keywords of the headline are utilized in forming the overall attention oriented representation. Secondly, by applying cardinal POS triplet and cardinal phrase guided attention, we ensure that cardinal value is emphasized and overall representation contains the effect of cardinal value.

%% file: 03-problem.tex

\section{Problem Definition and Proposed Model}

\begin{figure*}[!htbp]
      	\centering
       	 \vspace*{-5pt}
	\includegraphics[scale=0.3399]{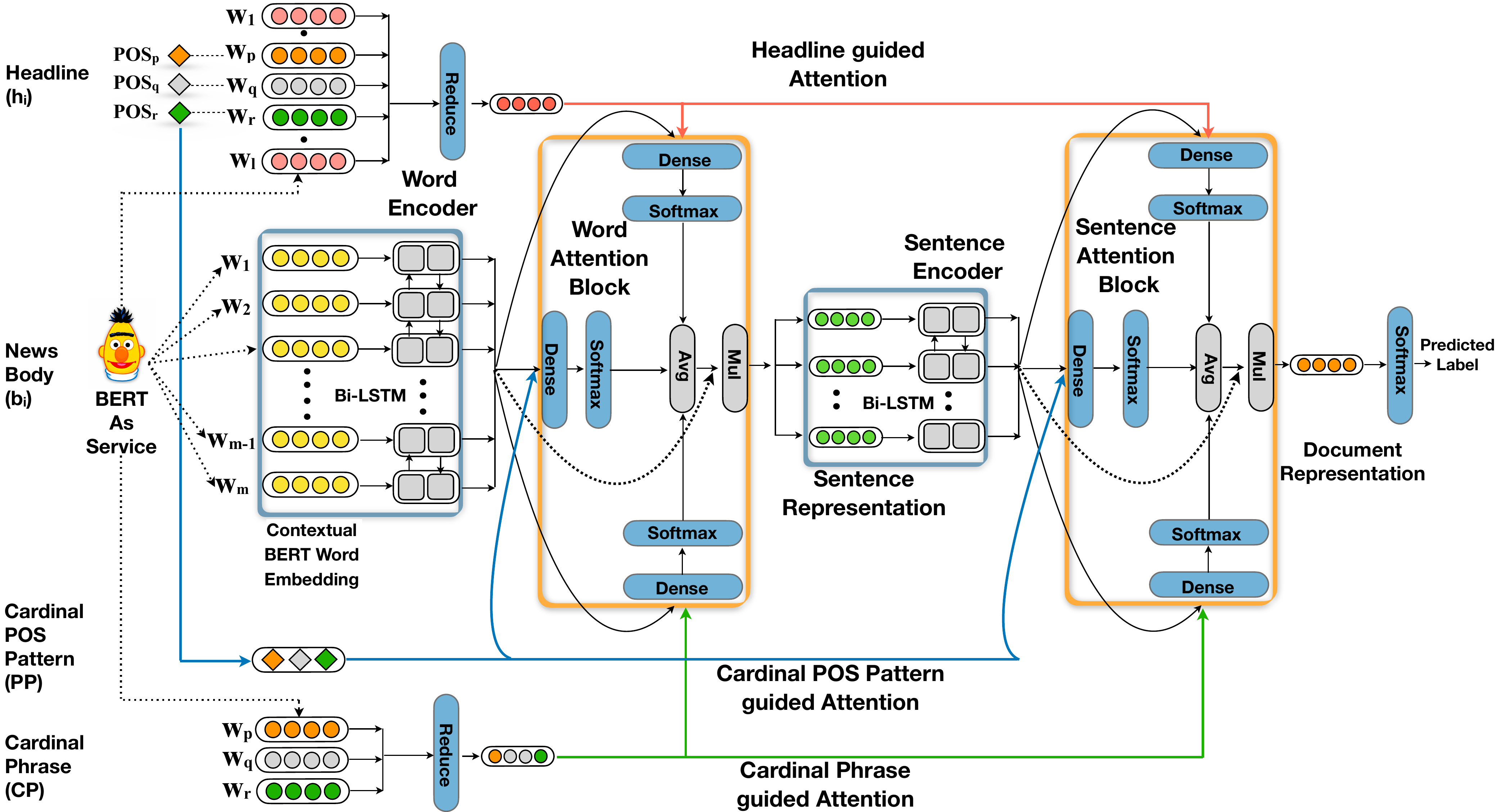}
      	\caption{Overall Architecture of POSHAN Model: Rectangle with blue borders are the Bi-LSTM based encoder at the word and sentence levels. Rectangle with orange borders are the Attention blocks at the word and sentence levels. Headline guided  Attention,Cardinal POS Pattern  guided Attention  and Cardinal Phrase  guided Attention are depicted as red, blue and green connecting lines respectively. }
      	\label{fig:Arch}
     \hspace*{5pt} 		 
\end{figure*}

\noindent
In this section, we formally introduce the problem definition.
Then we present the overall architecture of the proposed model \emph{POSHAN} sequentially, see Figure~\ref{fig:Arch}. In specific, we first discuss the embedding layer, which outputs the vector representations of the words and  cardinal pos-tag patterns. Secondly, we describe the cardinal pos-tag pattern guided hierarchical attention in detail for both word and sentence level. Thirdly, we introduce cardinal phrase guided and headline text guided hierarchical attention. In the end, we explain a method to fuse all the three attention types to get the overall attention scores. 
%

\subsection{Problem Definition}
Given a news item $ n_i\in N$, where $N$ is the set of all news items, which has a headline $h_i$ and body content $b_i$, \emph{news title incongruence detection} aims to predict the news as ``Congruent (C)'' or ``Incongreunt (I)', where incongruence denotes a mismatch between $h_i$ and $b_i$ by content. News headline $h_i$ and news body content $b_i$ are comprising of sequence of $l$ words as $h_i = \{w_{h1}, w_{h2}, ..., w_{hl}\} \in W$ and $m$ words as $b_i = \{w_{b1}, w_{b2}, ..., w_{bm}\} \in W$ correspondingly, where $W$ is the overall vocabulary set.  

\subsection{Embedding Layer}
\label{embedding}
\begin{figure}[t]
      	\centering
 	\hspace*{-8mm}
	\includegraphics[scale=0.27]{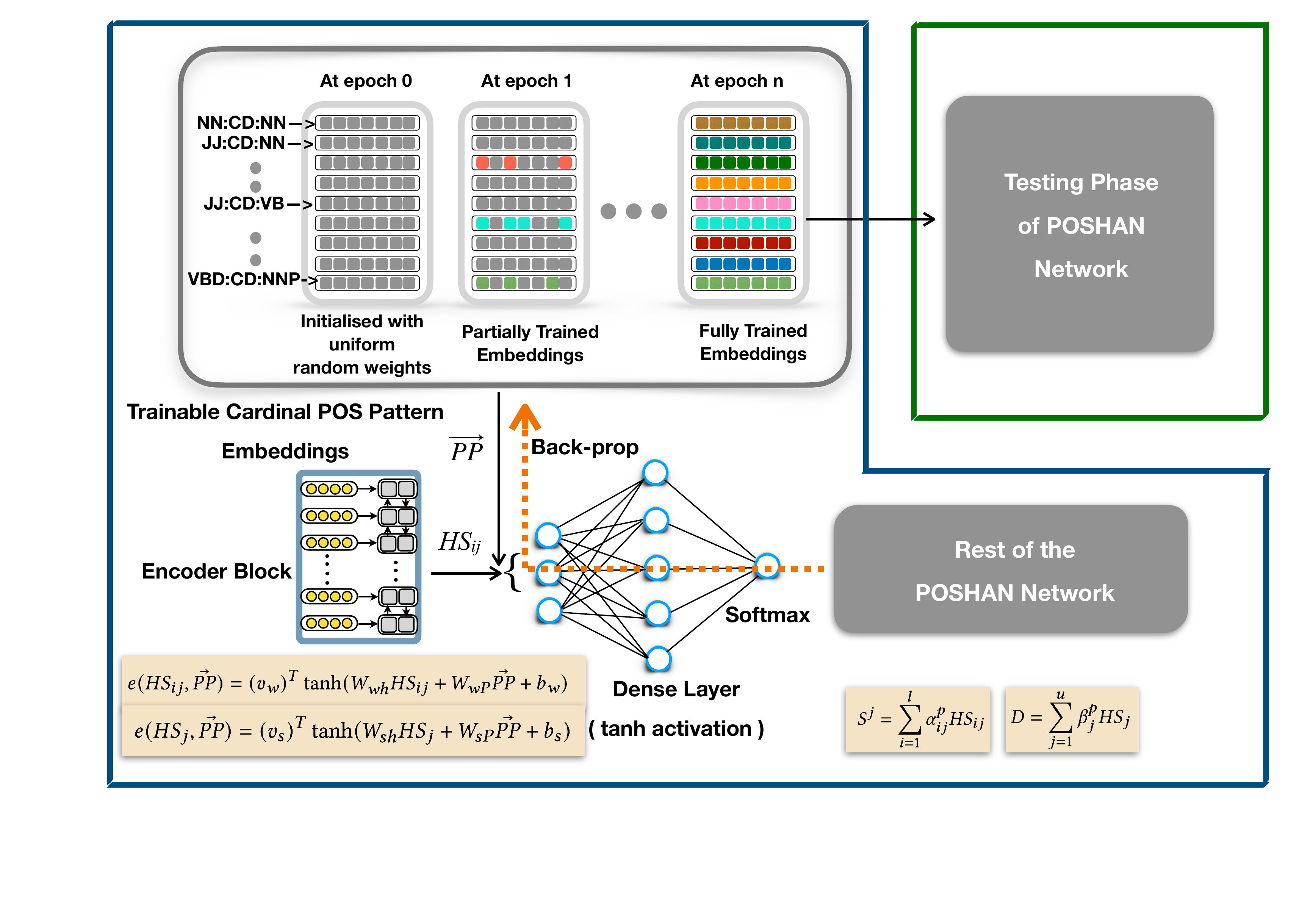}
	\vspace*{-30pt}
      	\caption{Training of Cardinal POS Pattern Embeddings: Inspired by a recent work \cite{sadhan}, we create 
trainable Cardinal POS Pattern embeddings of
100 dimensions for each POS pattern and initialize them with the uniformly random
weights to get the representation of POS patterns in vector space. The weights of these embeddings are trained during training of POSHAN model via back-prop of error. At the test time, we use already trained Cardinal POS Pattern embeddings, trained during training. 
}
      	\label{fig:Embedtrain}
      	\vspace*{-15pt}
\end{figure}

In Figure~\ref{fig:Arch}, for a news item $n_i$, the corresponding headline $h_i$ of length $l$ and body content $b_i$ of length $m$ are represented as $h_i = \{f(w_{h1}),..,f(w_{hl})\}$ where $f(w_{hj}) \in \mathbb{R}^d$ is a word embedding vector of dimension $d$ for $J^{th}$ word in headline $h_i $ and $b_i = \{f(w_{b1}), f(w_{b2}), ..., f(w_{bm})\}$ where $f(w_{bk}) \in \mathbb{R}^d$ is a word embedding vector of dimension $d$ for $K^{th}$ word in body content $b_i$. We use pre-trained contextual \textsc{BERT} embeddings, extracted using bert-as-service \cite{xiao2018bertservice} tool to get the embeddings of the size of 768 dimensions for each word. Each headline is associated with a cardinal pos-tag pattern of form $(POS_p \; POS_q \; POS_r)$, where $POS_p$, $POS_q$ and $POS_r$ are the pos-tags corresponding to the cardinal phrase $(W_p \; W_q \; W_r)$. We describe the cardinal pos-tag pattern and cardinal phrase in detail in sections~\ref{cardinal pos pattern} and \ref{cardinal phrase} respectively. We also create the vector representation $\vec{PP}$ for each of the cardinal pos-tag patterns of the size of 100 dimensions and initialize them with uniformly random weights.  We learn weights for these cardinal pos-tag pattern embeddings jointly in the \emph{POSHAN} model via backprop of error, as shown in the Figure~\ref{fig:Embedtrain}.%

%

\subsection{Sequence Encoder}\label{Sequence}
The pre-trained contextual BERT embeddings of the words of the news body text $b_i = \{f(w_{b1}), f(w_{b2}), ..., f(w_{bm})\}$ are fed to a Bi-directional Long short term memory (Bi-LSTM) unit \cite{Schmidhuber} based encoder, which encodes the news body text using standard LSTM equations. The output from Bi-LSTM units are the concatenations of forward and backward hidden states for each word, i.e., $HS_{i,j} = \overrightarrow{hs_{i,j}}\mathbin\Vert \overleftarrow{hs_{i,j}}$. Where $\overrightarrow{hs_{i,j}}$ and $\overleftarrow{hs_{i,j}}$ are the forward and backward hidden states of Bi-LSTM units. $HS_{i,j}$ is the overall hidden state for the $i^{th}$ word of the  $j^{th}$ sentence.


\subsection{\textbf{Cardinal POS Triplet Patterns}}
\label{cardinal pos pattern}
The idea of utilizing POS-tag Patterns to capture the intended context in natural language text is inspired by prior works \cite{justeson_katz_1995, potha}.
\citet{justeson_katz_1995} propose and utilize $7$ handcrafted part-of-speech (POS) patterns to extract significant and useful phrases from a long unstructured text. We utilize part-of-speech patterns containing cardinal POS tag `CD’ and call it cardinal POS triplet patterns. A cardinal POS triplet pattern can be defined as $(* :CD: *)$, where in place of wildcards, there can be $(JJ, NN, VB)$ etc., e.g. $(NN:CD:JJ)$. In contrast to \cite{justeson_katz_1995}, we do not handcraft a list of the viable POS patterns rather we use the all possible combination of POS patterns of length $3$, containing POS tag `CD’. We apply a neural attention layer in which, these cardinal POS triplets are used to guide the attention to select salient words and sentences which are significant for the POS pattern.

\subsection{\textbf{Cardinal POS Triplet Pattern Guided Hierarchical Attention}}
The objective of the Cardinal POS Triplet Pattern attention is to attend or select salient words that are significant and have some connotation with the cardinal phrase of the headline. Similarly, we aim to attend the salient sentences at the sentence level attention.
\citet{Yoon} have used headline guided attention to model the contextual representation of the news body text. However, we observe that the headline guided attention is not sufficient and effective, in case of headlines containing cardinal values. During experiments, we noticed that only headline-based attention convolutes the effective representation and fails to capture the influence of cardinal phrases on the overall document representation. 
We take a different and more logical design decision, in which we use part-of-speech patterns contained in each headline $h_i$ to guide the attention. We learn an embedding $\vec{PP}$ for each cardinal POS triplet pattern as discussed in section~\ref{embedding}.

\textbf{Computing Word Level Attention weights:}  
We use the embedding of cardinal POS triplet pattern $\vec{PP}$ to compute the attention scores given to each hidden state of the Bi-LSTM encoder.
\begin{equation}
\begin{aligned}
S^j = \sum_{i=1}^{l} \alpha^p_{ij}HS_{ij}  
\end{aligned}
\end{equation}
  Where $ HS_{ij}$ is the hidden state for the $i^{th}$ word of $j^{th}$ sentence and $l$ is maximum number of words in a sentence. $ \alpha^p_{ij}$ is the attention weight. $S^j$ is the formed sentence representation of $j^{th}$ sentence after attention scores are applied. The attention score $\alpha_{ij}$ can be defined as:

\begin{equation}
\begin{aligned}
 \alpha^p_{ij}=\frac{\exp(e(HS_{ij},\vec{PP}))}{\sum_{k=1}^{l}\exp(e(HS_{ik},\vec{PP}))}
 \end{aligned}
  \end{equation}
  Where $e$ is a $tanh$ based scoring function, which is used to compute the attention scores. $\vec{PP}$ is the POS-Tag pattern vector. The scoring function $e(HS_{ij},\vec{PP})$ can be defined as:
  \begin{equation}
\begin{aligned}
e(HS_{ij},\vec{PP}) =(v_w)^T\tanh(W_{wh}HS_{ij}+W_{wP}\vec{PP}+b_w )
\end{aligned}
\end{equation}
   Where  $v_w$ is weight vector at the word level. $W_{wh}$ and $W_{wP}$ are the weight matrices for hidden state and aspect vector and $b_w$ is bias at the word level respectively. 
 
\textbf{Computing Sentence Level Attention weights:} To compute sentence level  POS-Tag pattern driven attention weights, we use  POS-Tag pattern vector representation  $\vec{PP}$ and hidden states ${HS}^S_j$ from the sentence level BI-LSTM units as concatenations of both forward and backward hidden states ${HS}^S_{j} = \overrightarrow{{hs}^S_{j}}\mathbin\Vert \overleftarrow{{hs}^S_{j}}$ as follows:

\begin{equation}
\begin{aligned}
D = \sum_{j=1}^{o} \beta^p_{j}HS_{j}
\end{aligned}
\end{equation}
Where $HS_{j}$ is the hidden state for $j^{th}$ sentence and $\beta^p_{j}$ is the attention weight. $o$ is the maximum no of sentences in a news body text. $D$ is the formed document representation after the attention scores are applied. The attention score $\beta_{j}$ can be defined as:

\begin{equation}
\begin{aligned}
\beta^p_{j}=\frac{\exp(e(HS_{j},\vec{PP}))}{\sum_{k=1}^{o}\exp(e(HS_{k},\vec{PP}))}
\end{aligned}
\end{equation}
Where $e$ is a $tanh$ based scoring function, which is used to compute the attention scores. $\vec{PP}$ is the POS-Tag pattern vector. The scoring function $ e(HS_j,\vec{PP})$ can be defined as:
\begin{equation}
\begin{aligned}
e(HS_j,\vec{PP}) =(v_s)^T\tanh(W_{sh}HS_{j}+W_{sP}\vec{PP}+b_s )\end{aligned}
\end{equation}
Where $v_s$ is weight vector at the sentence level. $W_{sh}$ and $W_{sP}$ are the weight matrices for hidden state and aspect vector and $b_s$ is bias at the sentence level respectively.


\subsection{\textbf{Cardinal Phrase Guided Hierarchical Attention}}\label{cardinal phrase}
We deal with the incongruence detection for the news headlines containing cardinal numbers, therefore the most significant information and cue is the cardinal number itself and neighbouring words. For each headline, we extract a word triplet of form $*:Numerical-value: *$, where in place of wildcards, there can be any words., E.g. $Loan \: 1 \: million$. We call these word triplets as cardinal phrases. We use these cardinal phrases to drive attention to select salient words and sentences at word level and sentence level correspondingly.
To do that, we represent each cardinal phrase $CP$ as the summation of embeddings of all three words of word triplet as: 
\begin{equation}
\begin{aligned}
\vec{CP}= f(W_p) + f(W_q) + f(W_r)
\end{aligned}
\end{equation}

In a very similar fashion to cardinal POS triplet pattern guided attention, we use $\vec{CP}$ to compute the attention weights at both the word and sentence levels.
\begin{equation}
\begin{aligned}
 \alpha^c_{ij}=\frac{\exp(e(HS_{ij},\vec{CP}))}{\sum_{k=1}^{l}\exp(e(HS_{ik},\vec{CP}))} \text{~~\&~~} \beta^c_{j}=\frac{\exp(e(HS_{j},\vec{CP}))}{\sum_{k=1}^{o}\exp(e(HS_{k},\vec{CP}))}
 \end{aligned}
  \end{equation}
\subsection{\textbf{Headline Guided Hierarchical Attention}}\label{headline attention}
The objective of the headline driven attention is to select words and sentences in the news body text, which are relevant and topically aligned with headline content. The cardinal POS-tag pattern and cardinal phrase carry useful information regarding cardinal values but to capture the whole context of the headline and it's influence on news body text, we can not get rid of headline driven attention. 
We represent each headline $\vec{h}$ as the summation of embeddings of all the words contained in it as:\begin{equation}
\begin{aligned}
\vec{h} = \sum_{x=1}^{l}f(w_x)
\end{aligned}
\end{equation}
In a very similar fashion to cardinal POS triplet pattern guided attention, we use $\vec{h}$ to compute the attention weights at both the word and sentence levels.

\begin{equation}
\begin{aligned}
 \alpha^h_{ij}=\frac{\exp(e(HS_{ij},\vec{h}))}{\sum_{k=1}^{l}\exp(e(HS_{ik},\vec{h}))} \text{~~\&~~} \beta^h_{j}=\frac{\exp(e(HS_{j},\vec{h}))}{\sum_{k=1}^{o}\exp(e(HS_{k},\vec{h}))}
 \end{aligned}
  \end{equation}

                

\subsection{Fusion of Attention Weights and Classification}
\label{fuse}
We compute the overall attention weights from three kinds of attention mechanisms: POS-pattern-driven, Cardinal-phrase-driven, and headline driven attention at both the word and sentence levels. At the word level:
\begin{equation}
\begin{aligned}
\alpha_{i,j} = (\alpha^p_{i,j} + \alpha^c_{i,j} + \alpha^{h}_{i,j})/3 \text{~~\&~~} S^j = \sum_{i=1}^{l} \alpha_{ij}HS_{ij} 
\end{aligned}
\end{equation}
where $\alpha^p_{i,j} $, $\alpha^c_{i,j}$ and $\alpha^h_{i,j}$ are the attention weight vectors from POS-pattern, Cardinal-phrase and headline-attention at the word level. $S^j$ is the formed sentence representation after overall attention for the $j^{th}$ sentence. 
At the sentence level:
\begin{equation}
\begin{aligned}
\beta_j = (\beta^p_{j} + \beta^c_{j} + \beta^{h}_{j})/3 \text{~~\&~~} D = \sum_{j=1}^{o} \beta{j}HS_{j} 
\end{aligned}
\end{equation}
%
where $\beta^p_{j} $, $\beta^c_{j}$, and $\beta^h_{j}$ are the attention weight vectors from POS-pattern, Cardinal-phrase and  headline-attention at the sentence level, and $D$ is the formed document representation after overall attention. The document representation $D$ is used with a Softmax layer with softmax cross-entropy with logits as loss function for the classification. We compute the predicted label $\hat{y}$ as: $\hat{y} = softmax(W_{cl}D+ b_{cl})$.
Where $W_{cl}$ and $ b_{cl}$ are the weight matrix and bias term.

%% file: 04-data.tex
\section{Dataset Creation}

For evaluation, we create the datasets driven by two publicly available datasets,  NELA17 and Click-bait Challenge\footnote{\url{http://www.clickbait-challenge.org/}} (cf. Table~\ref{tab:data1}). 
\citet{Yoon} provide a script\footnote{\url{https://github.com/sugoiii/detecting-incongruity-dataset-gen}} to create the NELA17 dataset from an original news collection NELA17 dataset\footnote{\url{https://github.com/BenjaminDHorne/NELA2017-Dataset-v1}}. 
The NELA17 dataset comprises of 45521 congruent and 4551 incongruent news headline-body pairs.  The Click-bait Challenge dataset is created via crowd-sourcing based annotation of a collection of social media posts. 
The Click-bait Challenge dataset contains 16150 and 4883 social media posts, which are annotated as congruent and incongruent correspondingly.  
Using NELA17 and Click-bait Challenge datasets, we derive two new datasets, in which all the news headlines in NELA17 and all the social media posts in Click-bait Challenge, contain a numerical value. We call these two new datasets as Derived NELA17 dataset and Derived Click-bait Challenge dataset.
We create the new datasets using these steps:
\begin{enumerate}
\item We use POS tagger to get the words in headlines tagged with one of the corresponding Penn Treebank POS Tag Set.  
\item We keep all the headline-body pairs in which pos-tag CD (cardinal) appears in headline.
\end{enumerate}
The statistics of these datasets are reported in Table~\ref{tab:data2}.
We also extract two new features for each headline-body pair:
\begin{itemize}
\item A pos-tag triplet of form $* :CD: *$, where in place of stars, there can be JJ, NN, VB etc. E.g. $NN:CD:JJ$. We call this pos-tag triplet as Cardinal POS-tag Pattern. For a vector representation for each of the cardinal pos-tag pattern, we create a trainable embeddings of the size of 100 dimensions and initialize them with uniformly random weights. The weights for these embeddings are learned jointly using hierarchical attention in the \emph{POSHAN} model.
\item A word triplet of form $*:Numerical-value: *$, where in place of stars, there can be any words. E.g. $Loan \; 1 \; million$. We call this word triplet as Cardinal Phrase.
\end{itemize}

\begin{table}[t]
\caption{Dataset Statistics for NELA17}
\centering
        \begin{tabular}{p{2cm} p{2cm} p{3cm}}
                \toprule
            \bf Statistics    &\bf NELA17&\bf Derived NELA17 \\
            \midrule
                Incongruent &45521&6234\\
                   Congruent&45521&7766\\
                   Total &91042& 14000\\
\bottomrule
        \end{tabular}
 	\label{tab:data1}       
  	\vspace*{-5pt}
\end{table}
\begin{table}[t]
\caption{Dataset Statistics for Click-bait Challenge}
\centering
        \begin{tabular}{p{2cm}p{2cm}p{3cm}}
                \toprule
            \bf Statistics    &\bf Click-bait Challenge  &\bf Derived Click-bait Challenge\\
            \midrule
                Incongruent &4883&754\\
                   Congruent& 16150&2681\\
                   Total &21033&3435\\
\bottomrule
        \end{tabular}
 \label{tab:data2}       
 \vspace*{-5pt}
 \end{table}    

%% file: 06-results.tex

\section{Experimental Evaluation}
\subsection{Experimental Details}
\subsubsection{Baselines}
 We compare our model with the following baselines:
   \paragraph{\textbf{SVM}~\cite{cortes1995support}}: We start with feature-based methods utilizing support vector machine (SVM) by considering both linguistic and statistical features. 
   In specific, we use word tri-grams, four-grams, and part-of-speech bi-grams and tri-grams as features to learn a classifier using SVM. Usually, a click-bait headline contains word phrases like ``what happens if'' and ``You will Never Believe'', which can be easily captured by tri-grams and four-grams. Besides, POS tag combinations such as ``PRP WD RB'' are more frequent in incongruent headlines than incongruent ones, therefore, part-of-speech bi-grams and tri-grams are used to learn this distinguishing feature. 
   \paragraph{\textbf{LSTM}~\cite{Schmidhuber}}: We use long short term memory unit to encode both headline and body pair and apply softmax for the classification. We use pretrained GloVe embeddings of size 100 and the size of the hidden states of the Bi-LSTM unit is kept at 200. The concatenation of the news headline and the body text is used as input to the Bi-LSTM based encoder.
   \paragraph{\textbf{POSAt}}: 
   \label{sec:postag}
We propose a baseline method called as POS-tag guided Attention (POSAt) and compare the performance of our proposed approach \emph{POSHAN}, as this method uses POS tags to give importance to certain words. This method is inspired by a recent work \cite{postrip}. 
We use NLTK POS tagger to tag each word in the news headline/body pairs and maintain the mapping between words and corresponding POS tags using an index. POS tags are categorized into 6 semantic categories for sake of simplicity and brevity.
\begin{figure}[t]
      	\centering
	\includegraphics[scale=0.279]{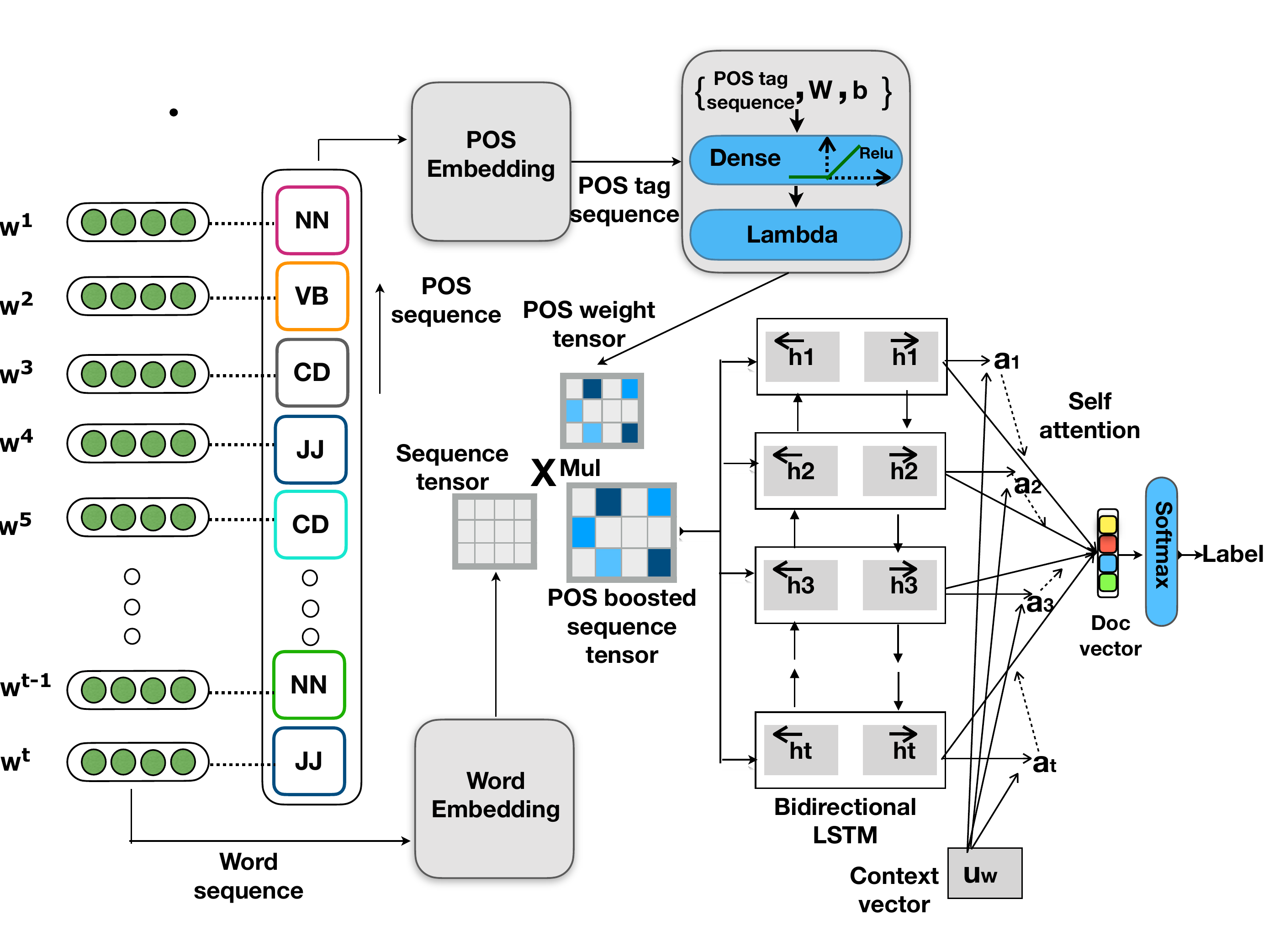}
      	\caption{Depiction of the POS tag guided attention model. From left to right, the sequence of words of the news item and corresponding part-of-speech tags go through word embedding and POS embedding layers. Attention scores are computed for each POS tag via a dense layer with ReLu activation. The resultant weight score matrix is multiplied to the word embedding sequence matrix. Attended representation is fed to a Bi-LSTM unit. Lastly, the resultant document representation from Bi-LSTM is used with a softmax layer for classification.}
      	\label{fig:postagatt}
\end{figure}
\noindent
\begin{enumerate}
\item \textbf{Noun chunk:} NN, NNS, NNP, NNPS
\item \textbf{Verb chunk:} VB, VBD, VBG, VBN, VBP, VBZ
\item \textbf{Adjective chunk:} JJ, JJR, JJS
\item \textbf{Pronoun chunk:} WP, WP
\item \textbf{Adverb chunk:} WRB
\item \textbf{Cardinal numbers chunk:} CD
\end{enumerate}
The POS tag for each word is represented as a 6-dimensional vector $g(x_i) \in \mathbb{R}^6$ of length 6. The weights of these embedding vectors are initialized in two ways.
\begin{enumerate}
\item \textbf{Initialize with very less value, close to zero:} In case of all zeros initialization, model performs well.
\item \textbf{Initialize with random weights:} In case of random weight initialization performance of the model degrades.
\end{enumerate}
These POS tag vector sequences are fed into a fully connected POS tag embeddings layer so that their weights are also trainable. Each part-of-speech category is assigned with one attention weight $\theta_i$, which will be learned
during training. In this way, each word is represented by$f'(w_i) = f(w_i)\times \theta_i$. 
We train a small neural network with only single hidden layer to learn weights for each POS tag category and then we use a custom lambda layer to reshape the POS weight tensor into a compatible shape so that we can boost each word vector with it's corresponding POS weight vector.

    In very similar fashion to LSTM baseline, pretrained GloVe embeddings of size 100 are used to represent the word vectors and hidden states of the LSTM unit is kept at 200. The nltk library\footnote{\url{https://pypi.org/project/nltk/}}with MaxEnt POS Tagger \cite{ratnaparkhi-1996-maximum} is used to tag the concatenation of the news headline and the body text.
     
   \paragraph{\textbf{Yoon}~\cite{Yoon}}: This is a state of the art method for news headline incongruence detection. It uses a hierarchical dual encoder based model which uses headline guided attention to learn the contextual representation. The original \cite{Yoon} paper uses a Korean news collection as dataset for evaluation but they also release an English version of the dataset called as NELA17. We use their model with NELA17 dataset, keeping all the settings as prescribed in \cite{Yoon}.
   \paragraph{\textbf{BERT-Sent\_Pair}~\cite{devlin-etal-2019-bert}}: We fine tune a pretrained BERT model for sequence pair classification task. We utilize the Hugging face transformers and dataset libraries to download pre-trained model. We use pre-built  "BertForSequenceClassification", provided by Hugging face library. Headlines and body pairs are packed together into a single sequence with adequate padding.
   \paragraph{\textbf{MuSem}~\cite{MuSem}}: This is a very recent work related to title incongruence detection, which uses both the NELA17 and Click-bait challenge datasets for evaluation. The authors propose a method that uses inter-mutual attention-based semantic matching between the original and a synthetically generated headlines via generative adversarial network based techniques, which utilises the difference between all the pairs of word embeddings of words involved and computes mutual attention score matrix.


\begin{table}[t!!!]
		\centering
		\caption{
		Comparison of the proposed model \emph{POSHAN} with various state-of-the-art baseline models for NELA17 Dataset. The results for \emph{POSHAN} are statistically significant ($p-value = 1.32e^{-2}$ for NELA17 Dataset using pairwise student's t-test}
 \centering
		\begin{minipage}{0.5\textwidth}	
		\centering
		\begin{tabular}{p{3cm}p{1.5cm}p{1.5cm}}
			
	\toprule
				\multicolumn{3}{c}{Derived NELA17 Dataset}\\
	\midrule		
				\bf Model & \bf  Macro F1 & \bf AUC. \\
\hline 

SVM \cite{cortes1995support} &0.608 &0.610\\
LSTM\cite{Schmidhuber}&0.627&0.639\\
POSAt&0.624&0.637\\
BERT-Sent\_Pair \cite{devlin-etal-2019-bert} & 0.642 & 0.658\\
Yoon \cite{Yoon}&0.653& 0.659\\
MuSeM\cite{MuSem}&0.703& 0.721\\
POSHAN &\textbf{0.748}& \textbf{0.763}\\
				  
		\end{tabular}
\label{tab:nelaresult}
	\end{minipage}
	
%
\begin{minipage}{0.5\textwidth}	
 \centering

			\begin{tabular}{p{3cm}p{1.5cm}p{1.5cm}}
	                \toprule
		
				\multicolumn{3}{c}{Original NELA17 Dataset}\\
	 \midrule			
				\bf Model & \bf  Macro F1 & \bf AUC. \\
 \midrule

SVM~\cite{cortes1995support} &0.622 &0.637\\
LSTM~\cite{Schmidhuber}&0.642&0.663\\
POSAt&0.648&0.669\\
BERT-Sent\_Pair\cite{devlin-etal-2019-bert} & 0.677 & 0.683\\
Yoon~\cite{Yoon} &0.685& 0.697\\
MuSeM\cite{MuSem}&0.752& 0.769\\
POSHAN &\textbf{0.765}& \textbf{0.783}\\

\bottomrule
		\end{tabular}
		\end{minipage}
 \vspace*{-10pt}
		\end{table}
		
		\begin{table}[htbp]
\begin{minipage}{0.5\textwidth}	
		\centering
		\caption{Comparison of the proposed model \emph{POSHAN} with various state-of-the-art baseline models for click-bait challenge dataset. The results for \emph{POSHAN} are statistically significant ($p-value = 2.29e^{-3}$ for click-bait challenge dataset using pairwise student's t-test.}
 \centering

					\begin{tabular}{p{3cm}p{1.5cm}p{1.5cm}}
			
				\toprule
				\multicolumn{3}{c}{Derived Click-bait challenge Dataset}\\
			\midrule	
				\bf Model & \bf  Macro F1 & \bf AUC. \\
 \midrule

SVM~\cite{cortes1995support} &0.596 &0.608\\
LSTM~\cite{Schmidhuber}&0.604&0.617\\
POSAt&0.614&0.620\\
BERT-Sent\_Pair\cite{devlin-etal-2019-bert} & 0.637 & 0.649\\
Yoon~\cite{Yoon} &0.646& 0.659\\
MuSeM\cite{MuSem}&0.698& 0.717\\
POSHAN &\textbf{0.739}& \textbf{0.748}\\
				  
		\end{tabular}
				\label{tab:CBCresult}
		\end{minipage}
		\begin{minipage}{0.5\textwidth}	

	\centering


					\begin{tabular}{p{3cm}p{1.5cm}p{1.5cm}}
			
		\toprule
				\multicolumn{3}{c}{Click-bait challenge Dataset}\\
	\midrule			
				\bf Model & \bf  Macro F1 & \bf AUC. \\ 
 \midrule
SVM~\cite{cortes1995support} &0.618 &0.629\\
LSTM~\cite{Schmidhuber}&0.630&0.641\\
POSAt&0.636&0.649\\
BERT-Sent\_Pair\cite{devlin-etal-2019-bert} & 0.653 & 0.662\\
Yoon~\cite{Yoon} &0.660&0.678 \\
MuSeM\cite{MuSem}&0.735& 0.747\\
POSHAN &\textbf{0.743}& \textbf{0.761}\\
				  
\bottomrule
		\end{tabular}
		\end{minipage}
 \vspace*{-15pt}		
		
		\end{table}

\subsection{ \emph{POSHAN} Implementation Details}\label{implement}
The \emph{POSHAN} model is implemented using \textsc{TensorFlow 1.10.0}\footnote{\url{https://www.tensorflow.org/install/source}} platform. For performance evaluation, Macro F1, and Area Under the ROC Curve (AUC) scores are used as performance
metrics. We keep the size of hidden states of bi-directional Long Short-term Units (LSTM) as 300, the size of embedding dimensions of pretrained \textsc{BERT}
\cite{devlin-etal-2019-bert} embeddings as 768. We use softmax cross-entropy with logits as the loss function. We keep the learning rate
as 0.003, batch size as 128, and gradient clipping as 6. The parameters
are tuned using a grid search. We use $50$ epochs for each model and apply
early stopping if validation loss does not change for more than 5 epochs. We
keep maximum words in a sentence as 45 and maximum number of sentences in a news body text as 35.

\paragraph{\textbf{Handling the multiple cardinal values}}: There are some cases where news headlines contain multiple cardinal values such as in fig \ref{fig:example}. At the training time, to utilize the context of all cardinal values present in the headline, we replicate the news headline and body pair for each cardinal value in train set. At the test time however, we concatenate all the learned cardinal POS tag vectors pertaining to the same news headline and use this overall POS tag vector to guide the attention.

\paragraph{\textbf{Extraction of BERT Embeddings}~\cite{xiao2018bertservice}}: We use bert-as-service, which utilizes extract\_features.py file from original BERT implementation, to extract the word embeddings from pretrained BERT model. We fine tune  uncased\_L-12\_H-768\_A-12 pretrained BERT model for sentence pair classification task. We set pooling\_strategy argument to NONE and use our own tokenizer. We use fine tuned BERT model to extract embeddings of 768 dimensions for each word.

\subsection{Results}
\label{results}
In this section, we compare the results of the \emph{POSHAN} model with the baselines and state-of-the-art methods. 

\subsubsection{Results for NELA17 Dataset}
\label{Results for NELA17}
In Table~\ref{tab:nelaresult}, we observe that in case of Derived NELA17 dataset, all the deep learning based methods outperform the non-deep learning method such as SVM model, which uses linguistic features and gets $0.608$  and $0.610$ in terms of Macro F1 and AUC. The POSAt model with Macro F1 score as $0.624$ and AUC as $0.637$, performs comparable with vanilla LSTM model. In our experiments, we introspect that the design decision in POSAt model to apply POS-tag guided attention at the POS-tag chunk level, does not result in effective representation and provides very less intended effects of POS types on words.
This way of POS-tag guided attention learns the attention score at the POS category level only as discussed in Section~\ref{sec:postag} such as Noun chunk, Verb chunk etc.

The BERT-Sent\_Pair with Macro F1 as $0.642$ and AUC as $0.658$, outperforms the POSAt model with significant difference. This gain can be attributed to the better contextual representation of words, learned in form of transformer based BERT embeddings. On the other hand, Yoon model \cite{Yoon} performs slightly better than BERT-Sent\_Pair with Macro F1 as $0.653$ and AUC as $0.659$. In addition to hierarchical encoder, which captures the complex structure of the news body content, having inherent hierarchical nature, Yoon model also uses a headline driven hierarchical attention, which not only selects salient and relevant words and sentences but also reduces the effective length of the news body.  In contrast, vanilla LSTM, POSAt and even BERT-Sent\_Pair model did not scale well for long text sequences. The MuSem model uses generative adversarial network based synthetic headline generation methods to generate a very low dimensional headline corresponding to news body and applies a novel mutual attention based semantic matching for incongruence detection. The MuSem model achieves significant gains over Yoon model, due to low dimensional representation of news body and effective semantic matching technique.
The proposed \emph{POSHAN} model beats all the other methods achieving  $0.748$ and $0.763$ as Macro F1 and AUC, respectively. The potential reason behind this better performance is superior document representation learned due to proposed attention mechanisms, which give adequate importance to significant cardinal values present in headline. In contrast, both Yoon and MuSem models fail to capture cues pertaining to cardinal patterns and phrases. 

In case of Original NELA17 Dataset, we notice a very similar trend as with derived NELA17 dataset, however performance of all the models improved with a significant margin. On the other hand, \emph{POSHAN} happens to yield more improvement in performance compared to other models for original dataset as a bonus. These gains can be attributed to cardinal POS-tag pattern based attention and cardinal phrase guided attention in addition to headline guided attention significantly.

\subsubsection{Results for Click-bait Challenge Dataset}
In case of both the derived and original click-bait challenge dataset also, we see a very similar performance chart. All the deep learning based methods outperform the non-deep learning method such as SVM model With  $0.596$ and $0.608$ in Macro F1 and AUC. The MuSem model with $0.698$ and $0.717$ in Macro F1 and AUC, outperforms all the other baselines. The proposed model \emph{POSHAN} performs better than MuSem model with significant gains and these gains can be explained by very similar reasoning, as provided in Section~\ref{Results for NELA17}.


\subsection{Ablation Study}

In Table~\ref{tab:ablation1}, we report an ablation study of \emph{POSHAN} using Derived NELA 17 Dataset. We used derived dataset for ablation study rather than original dataset because we want to asses the importance of different components of the \emph{POSHAN} with major focus of this paper, which is news headlines with important numerical values. In the ablation version 1), we remove the cardinal POS-tag pattern guided attention and keep the other two methods of attention intact and this step results in significant decrease in performance, which proves the usefulness and effectiveness of the cardinal POS-tag pattern guided attention. This corroborates with our original hypothesis and intuition. In the ablation version 2), we remove cardinal phrase attention and observe very similar decrease in performance. In the ablation version 3), we replace headline guided attention with headline encoder, in which we encode the words of news headlines in addition to news body words and concatenate the overall encoded sequence. We observe that without headline guided attention, model performs poorly because just a simple concatenation of encoded body and headline word sequences does not result in contextually important representation. We can conclude from 1), 2) and 3), that although all the three attention mechanism are effective individually too but combination of all the three becomes more effective. In the ablation version 4), we replace the pretrained BERT embbedings with GloVe \cite{NIPS2013_5021}, due to which the performance degrades drastically. The reason behind such a drop in the results is that the BERT embeddings provide superior contextual information than GloVe pretrained embeddings. We do not see much change in results in ablation version 5) as Bi-GRU \cite{cho-etal-2014-learning} and Bi-LSTM perform pretty much the same with our dataset. The performance of the model decreases a bit with replacement of Bi-LSTM with LSTM units in ablation version 6) and the obvious reason behind this better context learned by Bi-LSTM compared to LSTM units. 
		\begin{table}[t!!!]
\begin{minipage}{0.5\textwidth}	
		\centering
		\caption{Ablation study of POSHAN and Yoon\cite{Yoon} model conducted on derived NELA 17 Dataset.}
 \centering

			\begin{tabular}{p{5.6cm}p{1.3cm}p{.7cm}}
			
			\toprule
				\multicolumn{3}{c}{Derived NELA 17 Dataset}\\
		\midrule		
				\bf Scenario & \bf  Macro F1 & \bf AUC. \\
 \midrule
Original POSHAN &0.748 & 0.763\\
\midrule
1) Remove Cardinal POS Att&0.726&0.742\\
2) Remove Cardinal Phrase Att&0.731&0.749\\
3) Replace Headline Att with Headline Enc &0.648&0.669\\
4) Replace BERT with Glove &0.716 &0.736\\
5) Replace Bi-LSTM with Bi-GRU &0.746&0.761\\
6) Replace Bi-LSTM with LSTM &0.741&0.759\\

		\end{tabular}
\label{tab:ablation1}
\end{minipage}

\begin{minipage}{0.5\textwidth}	
		\centering
 \centering

			\begin{tabular}{p{5.6cm}p{1.3cm}p{.7cm}}
			
			\toprule
				\multicolumn{3}{c}{Derived NELA 17 Dataset}\\
	\midrule	
				\bf Scenario & \bf  Macro F1 & \bf AUC. \\
 \midrule
Original Yoon &0.653 &0.659\\
 \midrule
1) Remove Headline Att&0.593&0.595\\
2) Replace para to sent level Att&0.649&0.653\\
3) Replace Glove with W2V &0.610&0.618\\
4) Replace Bi-LSTM with Bi-GRU &0.652&0.657\\
5) Replace Bi-LSTM with LSTM &0.641&0.648\\

\bottomrule
		\end{tabular}
\label{tab:ablation2}	
	\end{minipage}	

			
		\end{table}		

\subsection{Error Analysis}

		\begin{table}[t!!!]
		\centering
		\caption{Error Analysis of Yoon model \cite{Yoon} and \emph{POSHAN} with Derived Click-bait challenge Dataset}
 \centering

			\begin{tabular}{p{1.5cm}p{2.5cm}p{2.5cm}}
			
			\toprule
				\bf Model & \bf  False Positives & \bf False Negatives \\
 \midrule
MuSem &201& 235\\
\emph{POSHAN}&179 &207\\
\bottomrule
	\end{tabular}
\label{tab:error}	
		\end{table}	
We conduct an error analysis of MuSem \cite{MuSem} and \emph{POSHAN} model with Derived Click-bait challenge dataset in Table~\ref{tab:error}. In the case of MuSem model, we observe 235 false negatives(FN) and 201 false positives (FP), on the other hand, \emph{POSHAN} produces 207 false negatives and 179 false positives. 
We notice that the major improvement with \emph{POSHAN} model, occurs in false negatives from 235 to 207, and most of these incorrectly predicted samples were related to important cardinal figures mentioned in the news headlines such as `Indiana couple admits to stealing 1.2 Million dollars from Amazon’. 
We also observe some incorrectly predicted false-positive cases by \emph{POSHAN} model because of the wrong POS-tag assignment by POS tagger and due to this \emph{POSHAN} misses out on the opportunity to consider those cardinal POS patterns and cardinal phrases.
\subsection{Visualization of Cardinal POS Pattern Embeddings}
		\begin{figure}[t!!!]
      	\centering
       	 \vspace*{-8pt}
 	
	\includegraphics[scale=0.21]{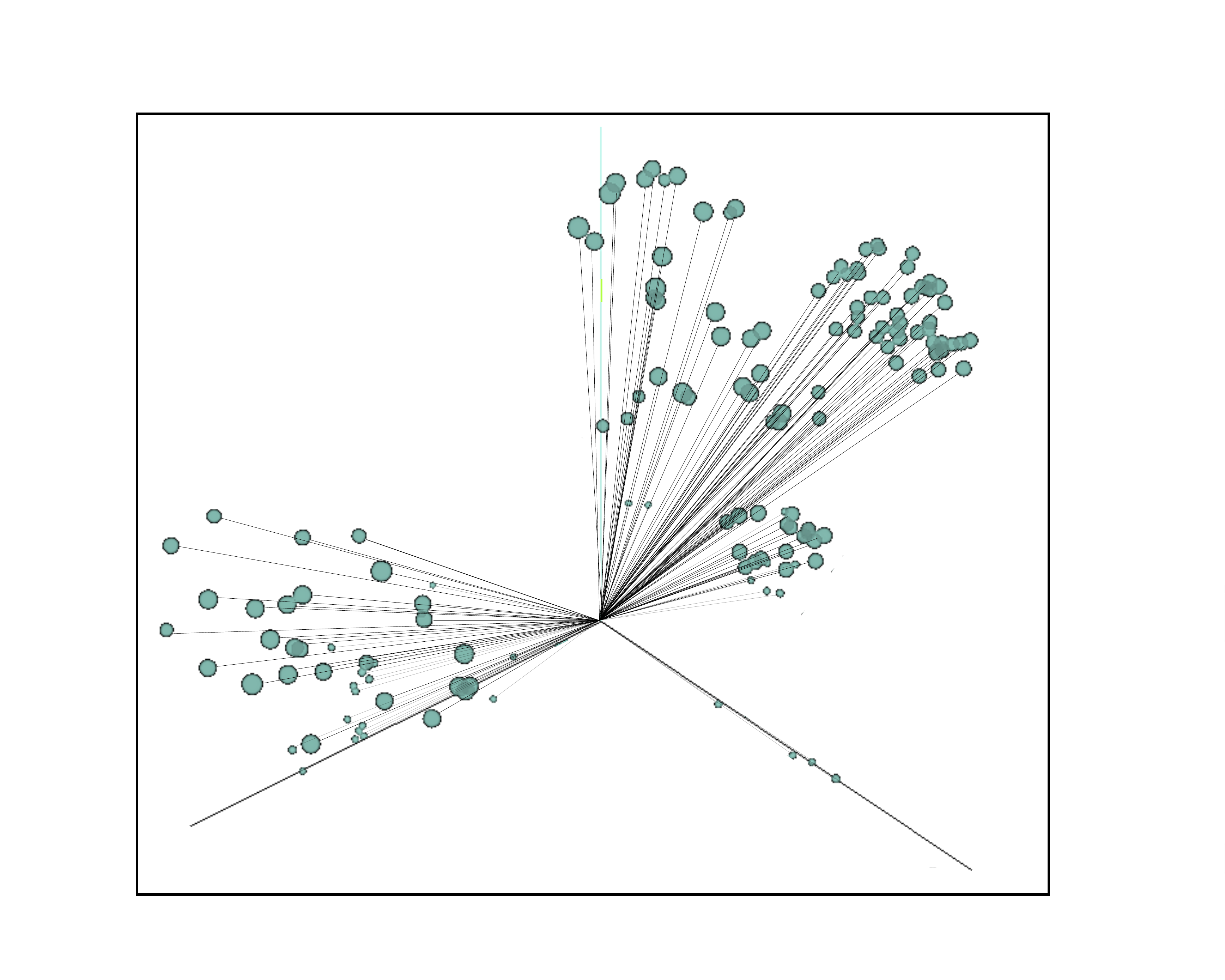}
	\vspace*{-12pt}
	\caption{Visualization  of Cardinal POS Pattern
Embeddings}      	\label{fig:embedViz}
      		  \vspace*{-10pt}
\end{figure}
In the Figure~\ref{fig:embedViz}, we present a visualization of cardinal pos-tag patterns. To visualize the learned embeddings of the cardinal pos-tag patterns, we use t-Distributed Stochastic Neighbor Embedding (t-SNE) \cite{Maaten08visualizingdata} with parameters as perplexity = 10, learning rate = 0.1 and iterations = 1000. The t-SNE method produces the visualization in a low dimensional space. We observe that the Cardinal POS Patterns have formed clearly separated clusters in embedding space, which connotes the congruence and incongruence labels. We also observe that cardinal POS patterns with similar tags such as $(NN:CD:JJ)$ and $(NNS:CD:JJ)$ are closer in embedding space and on the other hand the patterns with disjoint tag combinations such as $(NN:CD:JJ)$ and $(VBG:CD:CD)$ are farther apart from each other.

\subsection{Visualization of Attention Weights}
\begin{figure}[!htbp]
      	\centering
	\includegraphics[scale=0.21]{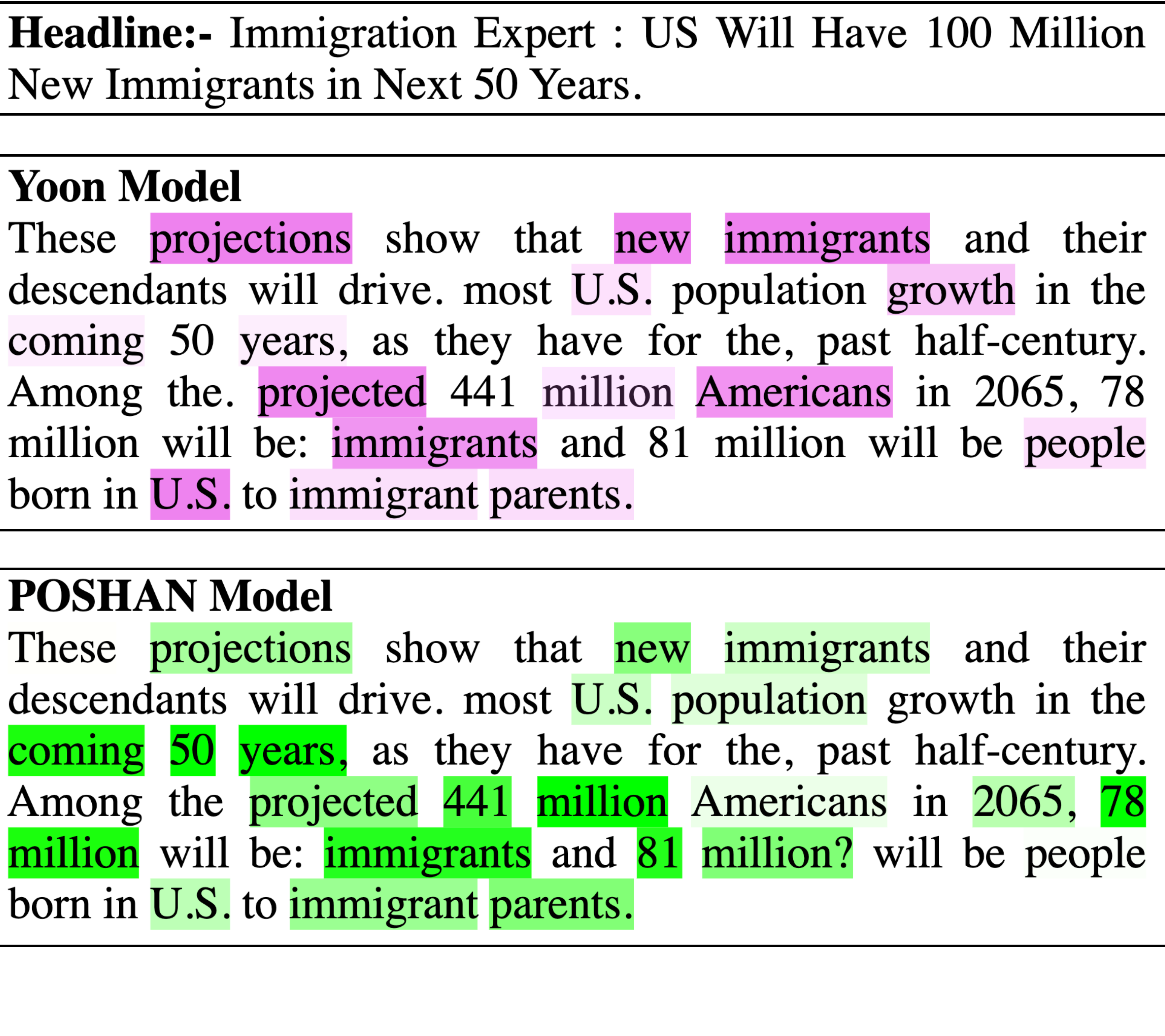}
 \vspace*{-10pt}
      	\caption{
      	Attention weight visualization: Word level attention weights from Yoon Model and POSHAN model for an anecdotal example are presented by highlighting the individual words (Best viewed in color). The depth of the color represents the strength of the attention weights.}
      	
      	\label{fig:attentionViz}
\end{figure}

In Figure~\ref{fig:attentionViz}, to analyse the interpretability of our model \emph{POSHAN} and to showcase the effectiveness of the proposed attention mechanism in forming the contextually important representations, we visualize the attention maps and compare it with Yoon model. In Figure~\ref{fig:attentionViz}, we use distribution of word level attention weights learned from both \emph{POSHAN} and Yoon model for an anecdotal example by highlighting the individual words. The depth of the color highlights represents the distribution of attention weights. Despite of common headline driven attention in both the models, we observe some clear differences between attention maps of Yoon model and \emph{POSHAN} model due to additional cardinal pos-tag pattern and cardinal phrase guided attention mechanisms in  \emph{POSHAN} model. The Yoon model successfully attends some words such as ‘immigrants’, ‘growth’ and ‘projections’ etc. relevant to headline context but fails to capture any words pertaining to significant cardinal phrases such as  ‘78 million’ and ‘50 years’ etc. On the other hand, \emph{POSHAN} model not only gives the importance to the words captured by Yoon model but also, it focuses on important cardinal phrases, which is in concert with our intuition about modeling the POS-tag pattern and cardinal phrase based attention.

%% file: 07-conclusion.tex
\section{Conclusions and Future Work}

In this paper, we introduce a novel task of incongruence detection in the news when the news headline contains cardinal values. The existing methods fare poorly as they fail to capture the context, pertaining to cardinal values. We present a joint neural model \emph{POSHAN}, which uses the fine-tuned BERT embeddings with three kinds of hierarchical attention mechanisms, namely cardinal POS-tag pattern guided, cardinal phrase guided and news headline guided attention. 
In the ablation study, we found that cardinal POS-tag pattern guided attention is very significant and effective in forming the cardinal quantity informed document representation. In the evaluation with two publicly available datasets, we notice that \emph{POSHAN} outperforms all the baselines and state-of-the-art methods. Visualization of cardinal POS-tag pattern embeddings and overall attention weights establish the effectiveness of the proposed model, decipher the model's decisions and make it more interpretable and transparent. In the future, we plan to model the degree of importance of cardinal values in news headlines and also we envisage an assessment of the applicability of the proposed model in case of textual entailment and fact verification tasks such as FEVER \cite{thorne-etal-2018-fever} dataset, in presence of cardinal values.